# A Comprehensive Survey on Multi-hop Machine Reading Comprehension Datasets and Metrics

Azade, Mohammadi

PhD Candidate, University of Isfahan, Isfahan, Iran, azade.mohammadi@eng.ui.ac.ir

Reza, Ramezani[*]

Assistant Professor, University of Isfahan, Isfahan, Iran, r.ramezani@eng.ui.ac.ir

Ahmad, Baraani

Professor of Computer Engineering, University of Isfahan, Isfahan, Iran, ahmadb@eng.ui.ac.ir

**Abstract**: Multi-hop Machine reading comprehension is a challenging task with aim of answering a question based on disjoint pieces of information across the different passages. The evaluation metrics and datasets are a vital part of multi-hop MRC because it is not possible to train and evaluate models without them, also, the proposed challenges by datasets often are an important motivation for improving the existing models. Due to increasing attention to this field, it is necessary and worth reviewing them in detail. This study aims to present a comprehensive survey on recent advances in multi-hop MRC evaluation metrics and datasets. In this regard, first, the multi-hop MRC problem definition will be presented, then the evaluation metrics based on their multi-hop aspect will be investigated. Also, 15 multi-hop datasets have been reviewed in detail from 2017 to 2022, and a comprehensive analysis has been prepared at the end. Finally, open issues in this field have been discussed.

**Keywords:** Multi-hop Machine Reading Comprehension, Multi-hop Machine Reading Comprehension Dataset, Natural Language Processing,

# 1- INTRODUCTION

Machine reading comprehension (MRC) is one of the most important and long-standing topics in Natural Language Processing (NLP). MRC provides a way to evaluate an NLP system's capability for natural language understanding. An MRC task, in brief, refers to the ability of a computer to read and understand natural language context and then find the answer to questions about that context. The emergence of large-scale single-document MRC datasets, such as SQuAD (Rajpurkar et al., 2016), CNN/Daily mail (Hermann et al., 2015), has led to increased attention to this topic and different models have been proposed to address the MRC problem, such as (D. Chen et al., 2016) (Dhingra et al., 2017)(Cui et al., 2017)(Shen et al., 2017).

However, for many of these datasets, it has been found that models don't need to comprehend and reason to answer a question. For example, Khashabi *et al* (Khashabi et al., 2016) proved that adversarial perturbation in candidate answers has a negative effect on the performance of the QA systems. Similarly, (Jia & Liang, 2017) showed that adding an adversarial sentence to the SQuAD (Rajpurkar et al., 2016) context will drop the result of many existing models. Also (D. Chen et al., 2016) pointed out that the required reasoning in the CNN/Daily Mail (Hermann et al., 2015) dataset is so simple that even a relatively simple algorithm can perform well on this dataset. Min et al. (Min et al., 2018) have shown that 90% of the questions in SQuAD (Rajpurkar et al., 2016), are answerable given only one sentence in a document, which does not involve complex reasoning.

The above problems seem to be due to the fact that answering the questions of these datasets doesn't require a deep understanding and reasoning (Khashabi et al., 2018). In other words, these datasets have focused only on answering questions based on a single or few nearby sentences of the context, mostly by matching information in the question and the context (known as single-hop MRC). However, in real-world cases, to answer a question it is required to read and comprehend multiple parts of disjoint evidence to find the valid information. Therefore, there are gaps between single-hop MRC, and real-world cases.

Multi-Hop Reading Comprehension (MHRC) is a more challenging extension of MRC in which it is needed to properly integrate multiple pieces of evidence and reason over them to correctly answer a question. In contrast to the question in single-hop MRC that can be answered by matching information, the multi-hop MRC task requires answering more complex questions based on a deep understanding of the full information. The reasoning ability is considered the first key in multi-hop MRC (Feng Gao et al., 2021).

Figure 1 shows two MRC triples. Figure 1(a) shows a single-hop question from (Rajpurkar et al., 2016) that as you can see this question can be answered with one sentence and mostly used matching information in the question and the context (*precipitation*, *fall*). Figure 1(b) shows an example from the WikiHop dataset (Song et al., 2020), where the machine has to gather information from three passages and reason over them to choose the correct answer among candidate answers. Passages contain two relevant facts: "*The Hanging Gardens are in Mumbai*" and "*Mumbai is a city in India*" and also some irrelevant facts, like: "*The Hanging Gardens provide sunset views over the Arabian Sea*" and "*The Arabian Sea is bounded by Pakistan and Iran*". As it is clear, finding the final answer in this kind of scenario is more challenging.

| | |
|---|---|
| In meteorology, precipitation is any product of the condensation of atmospheric water vapor that falls under gravity. The main forms of precipitation include drizzle, rain, sleet, snow, graupel and hail... Precipitation forms as smaller droplets coalesce via collision with other rain drops or ice crystals within a cloud. Short, intense periods of rain in scattered locations are called \showers". | [The Hanging Hardens], in [Mumbai], also known as Pherozshah Mehta Garden, are terraced garden… <br><br> [They] provide sunset views over [Arabian Sea].. <br> [Mumbai] (also known as Bombay), the official name until 1950 is the capital city of the Indian state of Maharashtra. [It] is the most populous city in [India]… <br><br> The [Arabian Sea] is a region and [Iran], on the west by northeastern [Somalia] and the Arabian Peninsula, and on the east by …. |
| Q: What causes precipitation to fall? | Q: (The Hanging Garden, Country, ?) <br> Options: Iran, India, Pakistan, Somalia, … |
| Answer: gravity | Answer: India |
| *(a)* | *(b)* |

*Figure 1: An example of single-hop and multi-hop question*

The first attempt to improve the simple single-hop MRC dataset happened with emerging of some datasets like TriviaQA (Joshi et al., 2017) and NarrativeQA (Kočisky et al., 2018). These datasets impose more challenges by introducing multiple passages per each question, and also presenting the questions that couldn't be answered with one single sentence. Although they present multiple passages per each question, still the most portion of the question in these datasets could be answered by exactly one sentence or a few nearby sentences within one passage, which means they do not need multi-hop reasoning for most percent of the questions. They are generally known as multi-passage or multi-document dataset that is closer to open-domain Question Answering or retrieving-reading problem, which means models have to focus on retrieving the most related passage and then answer the question based on that instead of reasoning over disjoint information. HotpotQA (Yang et al., 2018a) and WikiHop (Song et al., 2020) can be mentioned as the first and most popular multi-hop datasets which in addition to providing multiple passages per each question, ensure that the question can only be answered by reasoning over disjoint pieces of information. It has been shown that the models with successful results in single-hop MRC datasets have limited success on these datasets (Jiang & Bansal, 2019).

Recently, a lot of studies have been focused on different aspect of the multi-hop MRC task. Datasets are a vital part of the MRC task because, without a proper dataset, it is not possible to train and evaluate the models. It can be claimed that the emergence of datasets is the main motivation for much attention and progress in the MRC field after 2016. (Chen, 2018). Usually, the aim of each new dataset is to propose some new challenges that have been neglected by previous datasets, and then the existing models usually can't achieve the acceptable result in new datasets, and are needed to be improved. Therefore, it can be said that the high-quality multi-hop datasets promote the multi-hop models as well.

To have an accurate view of the growing trend of multi-hop dataset, the multi-hop models have to be considered as well (Figure 2). Several datasets with more complicated questions than single-hop datasets were introduced in 2017(like TriviaQA (Joshi et al., 2017)). Although they are not considered real multi-hop datasets, but they could attract models in 2018 to pay attention to multi-hop MRC task. Proposing multi-hop models has been beginning from 2018, but there was a shortage of real multi-hop datasets, so in 2018 more studies focused on creating multi-hop MRC datasets. These datasets made a proper situation to present the multi-hop MRC models, as you can see a significant number of models were been proposed in 2019. The trend of introducing new multi-hop datasets in 2020 to 2022 continued with 7 new datasets. It can be inferred from Figure 2 that whenever a new dataset has been created with new challenges, many studies focus on addressing those challenges, which shows the importance of the datasets.

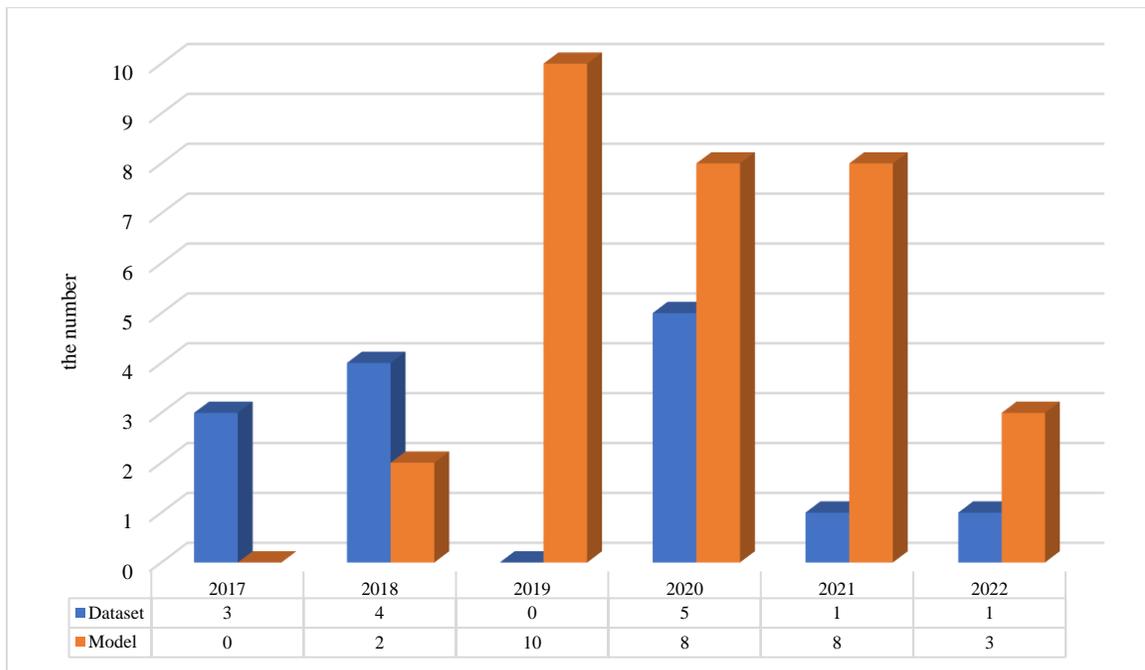

*Figure 2: The number of the multi-hop models and datasets in recent years*

There are some review papers on the MRC task. Liu et al. (Liu et al., 2019) reviewed 85 studies from 2015 to 2018 with a focus on neural network solutions for the MRC problem to investigate the neural network methods in the MRC task. Baradaran, Ghiasi and Amirkhani (Baradaran et al., 2020) presented a survey on the MRC field based on 124 reviewed papers from 2016 to 2018 with a focus on presenting a comprehensive survey on different aspects of machine reading comprehension systems, including their approaches, structures, input/outputs, and research novelties. Thayaparan, Valentino and Freitas (Thayaparan et al., 2020) proposed a systematic review about explainable MRC, from 2014 to 2020 with a focus on the explainable feature of the recent MRC methods. Zhang, Zhao and Wang (Zhang et al., 2020) presented a survey on the role of contextualized language models (CLMs) on MRC from 2015 to 2019. Bai and Wang (Bai & Daisy Zhe Wang, 2021) presented a survey on textual question answering with a focus on datasets and metrics, they investigate 47 datasets and 8 metrics. Also (Mavi, Jangra, and Jatowt 2022) presented a survey on multi-hop QA with 8 datasets and 29 models.

Although the mentioned surveys investigate different aspects of MRC/QA, none of them have focused on the multi-hop datasets and evaluation metrics. Due to increasing attention to multi-hop MRC, also the important role and the large number of recent studies on multi-hop datasets and evaluation metrics, it is necessary to investigate them separately. Our contribution in this paper is to propose a comprehension survey to investigate the existing studies on multi-hop evaluation metrics and datasets, their growth trend, and also the important challenges in this area. In this regard, we first introduce the problem definition of the multi-hop MRC task and investigate the existing evaluation metrics, then review 15 multi-hop MRC datasets from 2017 to 2022. Also, a fine-grain analysis of the datasets will be presented from different aspects. Finally, open issues in this field have been discussed.

It is important to note that since there is a close relationship between MRC and Question Answering, most of the existing machine reading comprehension tasks are in the form of textual question answering (Zeng et al., 2020), also MRC is known as a basic task of textual question answering (Liu et al., 2019). Figure 3 shows the relationship between QA, MRC and multi-hop MRC (Bai & Daisy Zhe Wang, 2021).

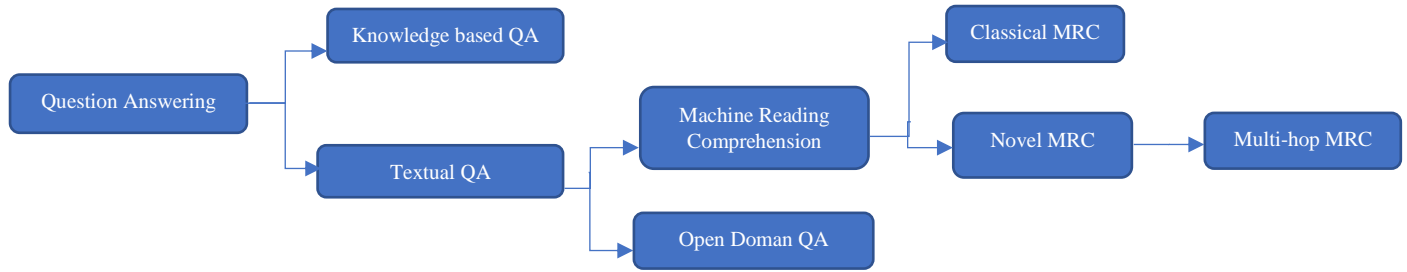

*Figure 3: The relationship between QA and MRC*

The rest of this paper is organized as follows: In Section 2, the definition of the problem and different types of the multi-hop MRC task are explained. In Section 3, evaluation metrics for multi-hop MCR are investigated, Next, in section 4, existing multi-hop MRC datasets are reviewed in detail and with a focus on their multi-hop aspects. Also, a fine-grain comparison of reviewed datasets will be presented in section 5. Open issues are expressed in Section 6, and finally Section 7 concludes the paper.

## 2- PROBLEM DEFINITION

In general, the multi-hop MRC problem can be defined as:

Given a collection of training examples $(C; Q; A)$, the goal is to find a function $F$ which takes a context $C$ and a corresponding question $Q$ as inputs, and gives answer $A$ as output.

$$F: (C, Q) \to A \quad (1)$$

Since the problem is multi-hop, $C = (P_1, P_2, \ldots, P_{l_p})$ can be a set of paragraphs (or documents) where $l_p$ denotes the number of paragraphs (or documents) and also question $Q$ is such a way that needs the multiple disjoint pieces of information from $C$ to be answered.

Like general MRC task, Answer $A$ in multi-hop MRC can be in different forms, where have been divided into four categories(D. Chen, 2018):

**Span-extraction:** The span extraction task needs to extract the subsequence $A$ from $P_i (P_i \in C)$ as the correct answer of question $Q$ by learning the function $F$, such that $A = F(C, Q)$.

**Multiple-choice:** Given a set of candidate answers $A = \{A_1, A_2, \ldots, A_n\}$, the multiple-choice task needs to select the correct answer $A_i$ from $n$ possible answer by learning the function $F$, such that $A_i = F(C, Q)$.

**Free-form:** The correct answer is $A$ that $A \subseteq C$ or $A \nsubseteq C$. In other words, the answer is not necessarily limited to be a part of the passage. The free-form task needs to predict the correct answer $A$ by learning the function $F$, such that $A = F(C, Q)$.

**Cloze-style:** The correct answer $A$ is part of the question $Q$ (usually a word or an entity) that is removed from question. The cloze style task needs to fill in the blank with the correct word or entity $A$ by learning the function $F$, such that $A = F(C, Q)$.

All General MRC tasks have the potential to be used in form of multi-hop MRC form as well. But as Figure 4 shows, most existing multi-hop datasets focus on the first two tasks (span-extraction and multiple-choice). it should be said that paying attention to the two first tasks is not specific for multi-hop and it happens for general MRC too. The great attention to these tasks may be because they are more natural and similar to the real-world case in comparison to the cloze-style task. Although the free-from task is more similar to the real-world case among all tasks, it didn't achieve proper attention which can be because of the complexity of this task for creating datasets and proposing models.

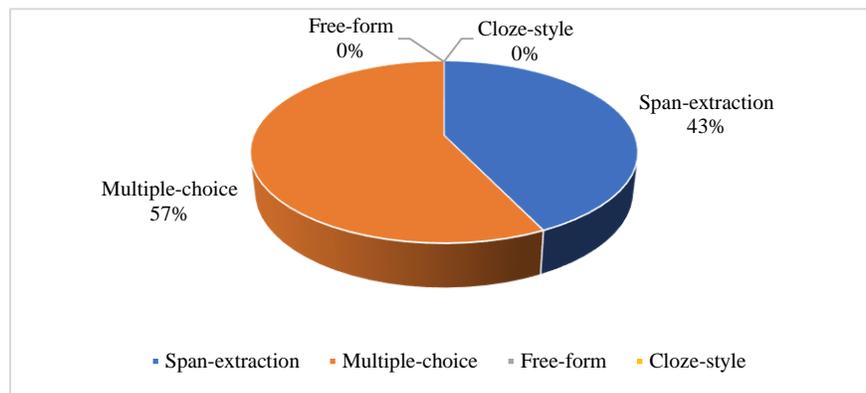

*Figure 4: The frequency of Multi-hop MRC tasks in reviewed datasets*

## 3- EVALUATION METRICS

Since evaluation metrics are an important aspect of each task, in this section, the evaluation metrics used in the multi-hop MRC task will be reviewed and investigated.

**Exact Match**: This metric is used to show whether two texts (predicated answer and the ground truth) are exactly the same. If they are exactly the same, EM will be 1, otherwise, it will be 0. This metric is a popular and useful metric in the span-extraction task, as in this task the predicated answer should be exactly matched with the ground-truth answer.

**F1 Score**: The F1 score metric measures the average overlap between the predicted answer and the ground-truth one which is widely used in the span-extraction task. It is calculated with precision and recall measurements, that we briefly show as follows:

$$Precision = \frac{TP}{TP+FP} \quad (2)$$

$$Recall = \frac{TP}{TP+FN} \quad (3)$$

$$F1 = \frac{2*Precision*Recall}{Precision+Recall} \quad (4)$$

where TP is True Positive, FP is False Positive, TN is True Negative and FN is False Negative. Consider the predicated and ground-truth answers as a bag of tokens. Accordingly, the definition of these parameter is shown in Table 1.

*Table 1: TP, FP, TN and FN (Liu et al., 2019)*

|  | **Tokens in ground-truth** | **Tokens not in ground-truth** |
|---|---|---|
| **Tokens in predicated answer** | TP | FP |
| **Tokens not in the predicated answer** | FN | TN |

Two previous metrics are common and popular in many NLP tasks as well as general MRC. There are three sets of metrics based on them that are more common and important in multi-hop MRC/QA models that have been introduced by Yang et al. (Yang et al., 2018a). The main goal of multi-hop MRC models is to find the answer then EM and F1 are used to evaluate this task known as the Answer task. Also, multi-hop MRC should present a series of evidence sentences that have led to the final answer known as the Support task as proof of multi-hop reasoning, this task is for evidence sentence extraction and could be seen as a kind of reasoning interpretability (Wu et al., 2021). Also, there is a task to evaluate both above tasks known as the Joint task. Each metric is explained in detail as follows:

- **Answer F1**: It focusses on the answer which is a span of the context. It measures the average overlap between the predicted answer span and the ground-truth one.
- **Answer EM**: It focuses on the answer which is a span of the context. It is used to show whether the predicated answer span and the ground-truth are exactly the same.
- **Support F1**: it focuses on the supporting fact and to show the expansibility of the Multi-hop MRC models. It measures the average overlap between of the supporting facts compared to the gold set.
- **Support EM**: It focuses on the supporting facts and to show the expansibility of the Multi-hop MRC models. It is used to show whether the supporting fact set and the gold set are exactly the same.
- **Joint F1**: it is to combine the evaluation of answer span and supporting facts: given precision and recall on the answer span ($P_{ans}, R_{ans}$) and the supporting facts $P_{sup}, R_{sup}$, respectively, joint F1 is calculated as (Yang et al., 2018a):

$$P^{joint} = P^{ans}P^{sup}, R^{joint} = R^{ans}R^{sup} \quad (5)$$

$$joint\ F1 = \frac{2P^{joint}R^{joint}}{P^{joint} + R^{joint}} \quad (6)$$

- **Joint EM**: is to combine the evaluation of answer span and supporting facts. It is 1 only if both tasks achieve an exact match and otherwise 0.

**Accuracy**: Accuracy is a popular and fairly common metric to evaluate the performance of the multiple-choice and cloze-style MRC tasks. In the multiple-choice task, it is required to check whether the correct answer has been selected from the candidate answers, and in the Cloze-style task, it is required to check whether the correct words have been selected for the missing words. The accuracy metric is calculated as follows:

$$Accuracy = \frac{N}{M} \quad (7)$$

where N is the number of questions answered correctly, and M is the number of all queried questions. Unlike EM and F1 score, there is no any supporting fact version of accuracy and it is only used for the answer prediction task.

**BLEU**: The BLEU (BiLingual Evaluation Understudy) metric was firstly proposed by Papineni et al. (Papineni et al., 2002) for machine translation, but it can be easily used in the free-form MRC task too. BLEU is used to calculate the similarity of two sentences. In the free-form MRC task, the predicted answer and the ground truth one is those two sentences whose similarity should be calculated. In this case, the probability of n-grams in the candidate sentence that appear in the ground truth will be calculated: (Liu et al., 2019)

$$p_n = \frac{\sum_i \sum_k \min(h_k(c_i), \max(h_k(r_i)))}{\sum_i \sum_k h_k(c_i)} \quad (8)$$

Where $h_k(c_i)$ counts the number of $k^{th}$ n-gram appearing in the candidate answer $c_i$. In a similar way, $h_k(r_i)$ denotes the occurrence count of that n-gram in the ground-truth answer $r_i$. If the length of the candidate sentence is very short, the accuracy of the BLEU score will decrease. To alleviate this problem, the penalty factor $BP$ is introduced, which can be calculated as follow (Liu et al., 2019):

$$BP = \begin{cases} 1, & c > r \\ e^{1-\frac{r}{c}}, & c \leq r \end{cases} \quad (9)$$

Finally, the BLEU is computed as:

$$BLEU = BP . \exp\left(\sum_{n=1}^{N} w_n \log p_n\right) \quad (10)$$

where $N$ means n-grams up to length $N$ and $w_n$ equals $1/N$. The BLEU score is the weighted average of each n-gram, and in most cases the maximum of $N$ is 4, namely BLEU-4.

**ROUGE-L**: ROUGE (Recall-Oriented Understudy for Gisting Evaluation) has been proposed by Lin (Lin, 2004) and is commonly used for the machine translation and summarization tasks, but it can be used for free-form MRC as well. L in Rouge-L is longest Common Subsequence (LCS). Rouge-L means applying the longest common subsequence to measure the similarity between text and can be calculated as follows (Liu et al., 2019):

$$R_{lcs} = \frac{LCS(X,Y)}{m} \quad (11)$$

$$P_{lcs} = \frac{LCS(X,Y)}{n} \quad (12)$$

$$F_{lcs} = \frac{(1+\beta)^2 R_{lcs} P_{lcs}}{R_{lcs} + \beta^2 P_{lcs}} \quad (13)$$

where $X$ is the ground-truth answer, m is the number of tokens of $X$, $Y$ is the predicated answer, n is the number of tokens of $Y$, $LCS(X,Y)$ is the length of the longest common subsequence of $X$ and $Y$, and $\beta$ is a parameter to control the importance of precision $P_{LCS}$ and recall $R_{LCS}$.

**Meteor**: Meteor (Metric for Evaluation of Translation with Explicit ORdering) has been proposed by Banerjeeand and Lavie (Banerjee & Lavie, 2005). Like two previous metrics, it was introduced for machine translation and can be used for free-form task as well. It is claimed that it has a high level of correlation with human judgments of translation quality in comparison to BLEU. It is calculated as follows:

$$Meteor = F * (1 - Penalty) \quad (14)$$

Where $F$ is a weighted F-score and is calculated as follows:

$$F = \frac{Precision * Recall}{\alpha * Precision + (1-\alpha) * Recall} \quad (15)$$

Also, $Penalty$ is calculated as follows:

$$Petalty = \gamma * \left(\frac{ch}{m}\right)^\beta \quad (16)$$

where $ch$ is the total number of matching chunks and m is the total number of matched uniforms between the prediction and the ground truth. The parameters $\alpha, \beta$, and $\gamma$ are tuned to maximize the correlation with human judgments.

All three metrics correspond with the way humans would evaluate the same text and also, and they are language-independent. They have a great potential to use in multi-hop MRC, but there are some weaknesses for them. The most important one is they do not consider the meaning of the words. For example, "watchman" and "guard" would consider two different words. Besides they do not consider the importance of the words in the corpus for scoring.

Figure 5 has summarized metrics and multi-hop MRC. Each metric has some features that make them suitable for one of the MRC tasks described in Section 2, then it cannot be said that one metric is better than the others as we cannot decide which one of the MRC tasks is better than the others. The frequency of usage of the evaluation metrics depends on the datasets. For example, Figure 4 showed that the number of the Span-extraction dataset is the most, then it can be concluded that EM and F1 scores are also the most used evaluation metrics which doesn't show their superiority, but it shows that there are more Span-prediction models and datasets.

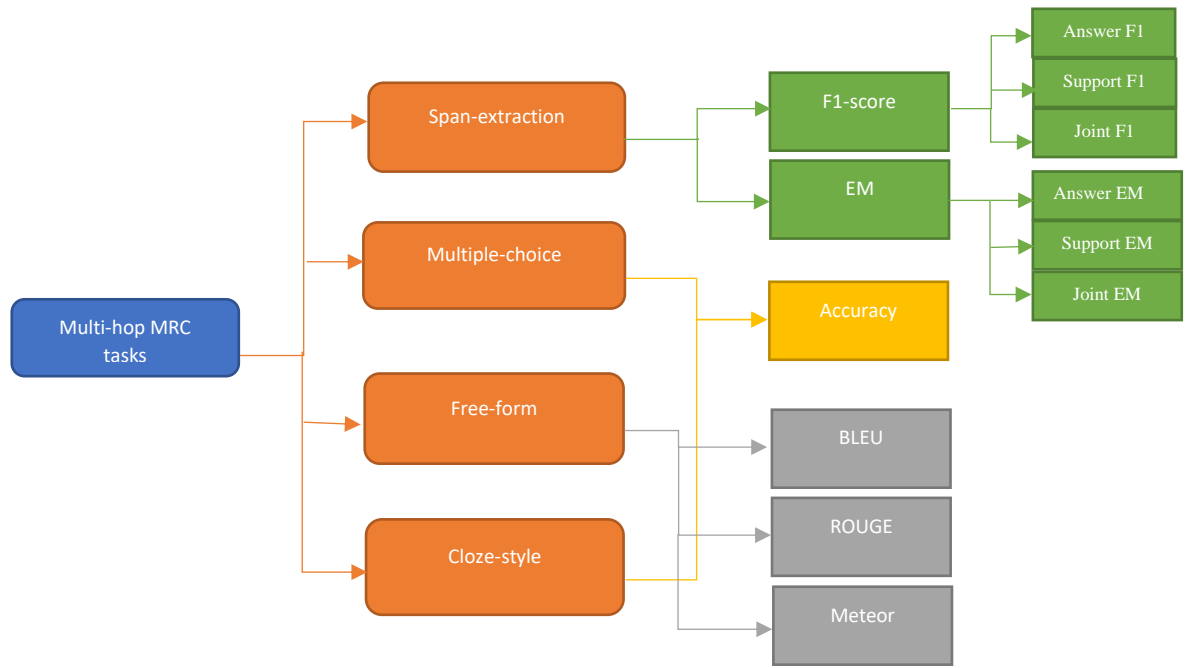

*Figure 5: The multi-hop MRC tasks and metrics*

## 4- DATASETS

In this section, 115 datasets have been reviewed in detail. We choose the datasets that focus on multi-hop challenges, which means it is required to gather information and reason on multiple disjoint pieces of information to answer the question.

Datasets are sorted mainly by release date, and we also start with the simplest datasets. These datasets introduced the question which cannot be answered with a single sentence and require gathering information from more than one sentence. Although they are not considered a multi-hop dataset today, due to the simplicity of the questions in comparison to recent multi-hop questions, they can be considered the first attempt to move from simple single-hop datasets (such as SQUAD (Rajpurkar et al., 2016)) to existing multi-hop datasets, then investigating them in this section will be useful. After a brief introduction of these simple datasets, we have investigated more complex multi-hop datasets that will be reviewed.

In the following, at first, each dataset will be introduced separately, in addition to general information, we focus on the points that make them suitable for multi-hop MRC. The end of section, contains a fine-grain comparison of reviewed datasets using several figures and tables.

**NewsQA** (Trischler et al., 2017) contains 119,633 plain-text questions on 12,744 news articles from CNN. The answers are spans that have to be extracted from context, but the length of spans is not fixed. This dataset also contains some questions that have no answer to examine the ability of the model to recognize inadequate information. The main goal of NewsQA is to emphasize reasoning behaviors. Table 2 shows the reasoning mechanism needed to answer questions in this dataset in comparison with the SQUAD dataset (Rajpurkar et al., 2016) which is a single-hop dataset. As you can see 20.7% of questions needs synthesizing information distributed across multiple sentences, and 13.2 % of questions need to be inferred from incomplete information or by recognizing conceptual overlap. However, in 27% of questions, a single sentence in the article entails the question, and 32.7% of questions can be answered with word matching. Although the number of complex questions (3 and 4) is more than SQUAD, in comparison to the number of the other type of questions (type 1 and 2) doesn't seem enough. besides, most of the questions in type 3 can be answered with a few sentences in one single paragraph and don't need complicated reasoning.

*Table 2: The reasoning mechanism (Trischler et al., 2017)*

|   | **Reasoning mechanism** | **NewsQA** | **SQUAD** |
|---|---|---|---|
| 1 | **Word matching** | 32.7 | 39.8 |
| 2 | **Paraphrasing** | 27 | 34.3 |
| 3 | **Inference** | 13.2 | 8.6 |
| 4 | **Synthesis** | 20.7 | 11.9 |
| 5 | **Ambiguous** | 6.4 | 5.4 |

**RACE** (Lai et al., 2017) has been collected from the English exams for middle and high school Chinese students and consists of 28,000 passages and 100,000 questions. Passages have different domains such as news, stories, ads, biography, philosophy, etc. The answers should be chosen from four candidate answers where only one of them is correct. One important point about this dataset is that the questions and answers are not limited to being from the original passage, they can be described in any words. RACE-M and RACE-H are two subgroups of RACE, where RACE-M denotes the Middle-school examinations and RACE-H denotes the High-school examinations. Some useful statistics of the RACE dataset are shown in Table 3. Also, Table 4 shows the type of reasoning mechanisms of this dataset in comparing two single-hop MRC datasets and also the NewsQA dataset (the mechanism the same as NewsQA). The number of questions from types 3 and 4 is remarkably more than CNN and SQUAD (two single-hop MRC datasets), and also is more than the NewsQA dataset, but still not enough portions of questions in RACE are in type 1(which means don't need reasoning over multiple disjoint information).

*Table 3: Statistics of the RACE dataset (Lai et al., 2017)*

| Feature | Dataset | | |
|---|---|---|---|
| | RACE-M | RACE-H | RACE |
| **Passage len.** | 231.1 | 353.1 | 321.9 |
| **Question len.** | 9.0 | 10.4 | 10.0 |
| **Option len.** | 3.9 | 5.8 | 5.3 |
| **Vocab size** | 32811 | 125120 | 132629 |

*Table 4: The reasoning mechanism (Lai et al., 2017)*

| | Reasoning mechanism | RACE | NewsQa | CNN/Daily mail | SQUAD |
|---|---|---|---|---|---|
| 1 | Word matching | 15.8 | 32.7 | 13 | 39.8 |
| 2 | paraphrasing | 19.2 | 27 | 41 | 34.3 |
| 3 | Inference | 33.4 | 13.2 | 19 | 8.6 |
| 4 | Synthesis | 25.8 | 20.7 | 2 | 11.9 |
| 5 | Ambiguous | 5.8 | 25 | 6.4 | 5.4 |

**TriviaQA** (Joshi et al., 2017) contains over 650K question-answer-evidence tuples, the context consists of the web documents and wiki documents while the older dataset like SQUAD(Rajpurkar et al., 2016) is limited to wiki documents and the News is limited to the news article. Some general statistics of TriviaQA are shown in Table 5. One of the advantages of this dataset is the syntactic and lexical variability between questions, answers, and context sentences which makes the reasoning difficult. Besides, there are six documents per question and also 40% of the questions need reasoning over multiple sentences, which is about three times more than SQUAD, and also is more than NewsQA (Trischler et al., 2017) and RACE (Lai et al., 2017). Figure 6 shows the percentage of the multi-sentences question in 5 datasets. However, most of these questions can be answered by a few nearby sentences in one single paragraph, and does not require complex reasoning (Yang et al., 2018b). But this dataset is still (2021) used by some multi-hop MRC models that shows the importance of this dataset.

*Table 5: TriviaQA statistics (Joshi et al., 2017)*

| Total number of QA pairs | 95,956 |
|---|---|
| **Number of unique answers** | 40,478 |
| **Number of evidence documents** | 662,659 |
| **Avg. question length (word)** | 14 |
| **Avg. document length (word)** | 2,895 |

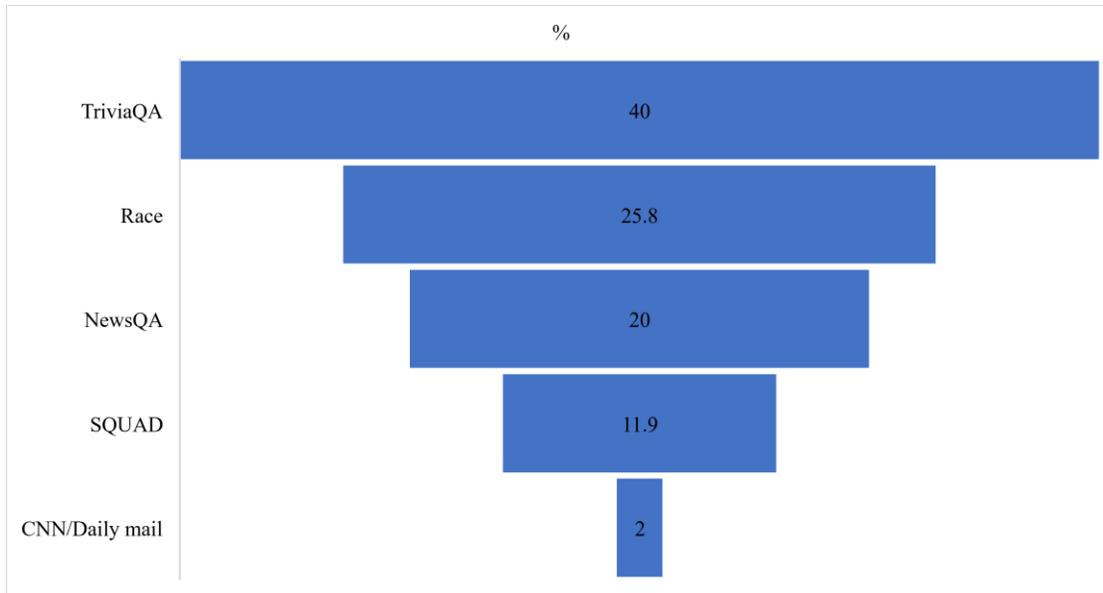

*Figure 6: The percentage of the multi-sentences Questions*

**COMPLEXWEBQUESTIONS** (Talmor & Berant, 2018) has been built from the WEBQUESTIONS (Berant et al., 2013) and includes 34,689 question-answer pairs. This dataset is important because generating complex questions that are needed for reasoning over multiple pieces of information is the main aim challenge of this dataset, and it is claimed that all of the questions are in this form. As it has been shown in Figure 7, to generate complex questions, they used the dataset WEBQUESTIONSSP (Yih et al., 2016), which contains 4,737 questions paired with SPARQL queries for Freebase (Bollacker et al., 2008), but since the questions are simple, they first generated automatically questions that are understandable to AMT workers, and then have them paraphrase those into natural language. They make sure that questions include phenomena such as composition questions (45%), conjunctions (45%), superlatives (5%), and comparatives (5%). A drawback of this method for question-generating is that because queries are generated automatically the question distribution is artificial from a semantic perspective.

| Seed Question | What movies have Robert Pattinson starred in? |
|---|---|
| SPARQL | ns:Robert_Pattinson ns:fil.actor.fils ?c |
|  | ?c ns:film.performance.film ?x . |
|  | ?x ns:film.produced_by ns:Erwin_stoff |
| Machine-generated | What movies have Robert Pattinson starred in and that was produced by Erwin Stoff? |
| Natural Language | Which Robert Pattinson film was produced by Erwin Stoff? |

*Figure 7: Data collection process (Talmor & Berant, 2018)*

**OpenBookQA** (Mihaylov et al., 2018) consists of 6000 plain-text questions based on 1326 elementary-level science facts. The answers are plain text which should be chosen from multiple candidate answers. Unlike other datasets that are self-contained answers, an open book fact and broad common knowledge are used together to answer the questions they require multi-hop reasoning over core facts and common knowledge to answer the question. However, it should be said that reasoning over a whole document is rarely needed actually, and also it is unclear how many additional facts are needed (Khot et al., 2020). Some usefull statistics of OpenBookQA are shown in Table 6.

Table 6: Statistics of the OpenBookQA dataset (Mihaylov et al., 2018)

| | |
|---|---|
| # of questions | 5957 |
| # of choices per question | 4 |
| Avg. question sentences | 1.08(6) |
| Avg. question tokens | 11.46(76) |
| Avg. choice tokens | 2.89(23) |
| Avg. science fact tokens | 9.38(28) |
| Vocabulary size (q+c) | 11855 |
| Vocabulary (q+c+f) | 12839 |
| Answer is the longest choice | 1108(18.6%) |
| Answer is the shortest chioce | 216(3.6%) |

**MultiRC** (Khashabi et al., 2018) consists of ~9k high-quality multiple-choice RC questions. It has been ensured that answering each of the questions requires reasoning over multiple sentences. This dataset is constructed using 7 different domains (news, Wikipedia articles, Articles on society, law and justice, Articles on history and anthropology, Elementary school science textbooks, 9/11 reports, and Fiction). You can see the statistics of this dataset in Table 7. Answering 60% of the questions of this dataset requires reasoning over multiple sentences. also, the required information to answer the questions is not explicitly stated in the context but is only inferable from it (e.g., implied counts, sentiments, and relationships). The number of multi-sentence questions is 5825 (from 9,872) and the number of sentences that are used to answer a question is 2.58 on average, these sentences are not continuous and the average distance between them is 2.4. However, this dataset has focused on passage discourse and entity tracking, rather than relation composition (Khot et al., 2020).

Table 7: Statistical information of MultiRC (Khashabi et al., 2018)

| Parameter | Values |
|---|---|
| **Number of Paragraphs** | 871 |
| **Number of Questions** | 9,872 |
| **Number of multi-sentence questions** | 5,825 |
| **Average number of candidates (per question)** | 5.44 |
| **Average number of correct answers (per question)** | 2.58 |
| **Average number of sentences used for questions** | 2.37 |
| **Average distance between the sentences used for each question** | 2.4 |

**QAngaroo** (Welbl et al., 2018) is composed of two multi-hop reading comprehension datasets: MedHop which is a cloze-style dataset, and WikiHop which is a multiple-choice dataset. The MedHop dataset is about molecular biology and consists of 1.6K instances for training, 342 instances for development, and 546 instances for testing. WikiHOP dataset consists of 51k questions, answers, and context where each context consists of several documents from Wikipedia. The size of WikiHop and MedHop datasets for different learning phases are shown in Table 8. In these datasets, models have to combine evidence across multiple documents and perform multi-hop inference. WikiHop is more popular Which is probably due to being open-domain and also the number of its questions. Each question in WikiHop is a tuple, which denotes two entities, and their relationship, then the answers in the WikiHop dataset are a single entity. As Table 9 shows, for 45% of cases, the answer can be found from multiple contexts, and for 9% of the cases, a single document suffices. Also, for 26% of cases, more than one candidate is plausible, (this is often due to hypernymy).

However, the questions of the WikiHop dataset are not in natural text form, which makes it different from real-world cases. Also, it is created by documents (from Web or Wikipedia) and a knowledge base (KB), as a result, these datasets are constrained by the schema of the KBs they use, and therefore the diversity of questions and answers is inherently limited (Yang et al., 2018a). also, these datasets have no information to explain the predicted answers (Ho et al., 2020).

*Table 8: Dataset sizes for the WikiHop and MedHop datasets (Welbl et al., 2018)*

|  | Train | Dev | Test | Total |
|---|---|---|---|---|
| **WikiHop** | 43738 | 5129 | 2451 | 51318 |
| **MEDHOP** | 1620 | 342 | 546 | 2508 |

*Table 9: Type of the questions in WikiHop (Welbl et al., 2018)*

| Type of question | Percentage (%) |
|---|---|
| multi-step answer | 45 |
| Multiple plausible answers | 15 |
| Ambiguity due to hypernymy | 11 |
| Only single document required | 9 |

**HotpotQA** (Yang et al., 2018a) has become one of the most popular multi-hop MRC datasets in recent years. It has 113k Wikipedia-based question-answer pairs. To answer the questions of this dataset, it is required to collect and reason over information from multiple supporting documents. In contrast to WikiHop (Welbl et al., 2018), the question and answer are in plain text form. The answers are variable-length spans, and there are 10 wiki passages per question. One of the distinguishing features of this database is that there are sentence-level supporting facts for each answer which is the sentences containing information that supports the answer. Thus, models should extract a span as the answer and also provide the supporting facts for the answers. The types of answers are shown in Table 10.

However, there are some drawbacks to this dataset, for example, QASC(Khot et al., 2020) discussed that since the questions of this dataset were authored in a similar way, due to their domain and task setup, they are easy to decompose, Also, as discussed in Inoue et al (Inoue, 2020), the task of classifying sentence-level SFs is a binary classification task that is incapable of evaluating the reasoning and inference skills of the model. Besides, (Chen and Durrett, 2019; Min et al., 2019) revealed that many examples in this dataset do not require multi-hop reasoning to solve (Ho et al., 2020).

*Table 10: Types of answers in the HotpotQA dataset (Yang et al., 2018a)*

| Answer type | % | Example(s) |
|---|---|---|
| **Person** | 30 | King Edward п, Rihanna |
| **Group / Org** | 13 | Cartoonito, Apalachee |
| **Location** | 10 | Fort Richardson, California |
| **Date** | 9 | 10th or even 13th century |
| **Number** | 8 | 79.92 million, 17 |
| **Artwork** | 8 | Die schweigsame Frau |
| **Yes/No** | 6 | - |
| **Adjective** | 4 | Conservative |
| **Event** | 1 | Prix Benois de la Dance |
| **Other** | 6 | Cold War, Laban Movement |
| **Noun** |  | Analysis |
| **Common noun** | 5 | Comedy, both men and women |

**R⁴C** (Inoue, 2020) is proposed to evaluate the internal reasoning of the reading comprehension system in HotpotQA (Yang et al., 2018a). As it's said before, HotpotQA requires identifying supporting facts (SFs), but since only a subset of SFs contributes to the necessary reasoning, thus, achieving high accuracy performance in the Support task (it has been explained in section ) cannot prove a model's reasoning ability. R⁴C contains 4,588 questions and aims to evaluate a model's internal reasoning in a finer-grained manner than the Support task in HotPotQA. It requires giving not only answers but also derivations. A derivation is a semi-structured natural language form that is used to explain and justify the predicted answers. Each question is annotated with 3 reference derivations (i.e., 13.8k derivations). However, the small size of the dataset is a serious limitation to train and evaluate end-to-end systems (Ho et al., 2020). Table 11 shows the statistics of the $R^4C$ corpus. "st." denotes the number of derivation steps. Each instance is annotated with 3 golden derivations.

*Table 11: The number of derivations in R4C (Inoue, 2020)*

| Split | QA# | #Derivations | | | |
|---|---|---|---|---|---|
| | | 2st | 3st | >4st. | Total |
| **Train** | 2,379 | 4944 | 1,553 | 640 | 7,137 |
| **Dev** | 2,209 | 4,424 | 1,599 | 604 | 6,627 |
| **Total** | 4,588 | 9,368 | 3,152 | 1,244 | 13,764 |

**2WikiMultiHopQA** (Ho et al., 2020) is a multi-hop dataset based on Wikipedia and WikiData[1], and consists of 192k multi-hop questions. An important point about this dataset is that it contains both structured and unstructured data. There are four types of questions in this dataset: 1) Comparison questions which are used to compare two or more entities, 2) Inference questions to infer relation r from the two relations r1 and r2, 3) Compositional questions to achieve the answer from an entity and two relations without any inference relations, 4) Bridge comparison which requires finding bridge entities. each sample contains some information, namely evidence that is used to explain the answer. Evidence is in from of triples, where each triple is structured data (subject entity, property, object entity) obtained from WikiData. You can see the number of examples in each type along with the average length of questions and answers of 2WikiMultiHopQA dataset in Table 12. This dataset like HotPotQA (Yang et al., 2018a) uses the Answer, Support and Joint evaluation metrics.

*Table 12: Question and answer lengths across the frequency questions (Ho et al., 2020)*

| Type of question | Number of Examples | Avg. length of question | Avg. length of answer |
|---|---|---|---|
| **Comparison** | 57,989 | 11.98 | 1.58 |
| **Inference** | 7,478 | 8.41 | 3.15 |
| **Compositional** | 86,979 | 11.43 | 2.05 |
| **Bridge-comparison** | 40,160 | 17.01 | 2.01 |
| **Total** | 192,606 | 12.46 | 1.94 |

**HybridQA** (W. Chen et al., 2020) is a large-scale multi-hop QA dataset with heterogeneous data based on Wikipedia tables and their relevant texts and hyperlinks, then to answer the questions, tabular and textual information should be combined. The dataset consists of 70K question-answer pairs with 13K Wikipedia tables. To generate this database, some HITs (human intelligence tasks) have been defined. In each HIT, a table and its content have been presented to a crowd worker to create 6 questions and their corresponding answers. Tables have been limited to tables with 5 to 20 rows and 3 to 6 columns. For each hyperlink in the table, the first 12 sentences of the Wikipedia page have been retrieved. The types of multi-hop reasoning used in this dataset are as follows: 1) Table → Passage chain: it first finds the cells in the table, and then hop to their neighboring hyperlinked cells within the same row, finally extracts a text span from the passage of the hyperlinked cell as the answer. 2) Passage → Table chain: it first finds the related passage, which traces back to certain hyperlinked cells in the table, and then hop to a neighboring cell within the same row, finally extracts text span from that cell. 3) Passage → Table →Passage chain: it is similar to Type 2, but in the last step, it hops to a hyperlinked cell and extracts the answer from its linked passage. 4)Passage and Table jointly are used to identify a hyperlinked cell based on table operations and passage similarity and then extract the plain text from that cell as the answer. 5) involves two parallel reasoning chains, while the comparison is involved in the intermediate step to find the answer. 6)involve multiple reasoning chains, while superlative is involved in the intermediate step to obtain the correct answer. The percentage of each type of reasoning has been shown in Table 13.

---

[1] https://www.wikidata.org/wiki/Wikidata:Main_Page

Table 13: The percentage of each type of reasoning (Chen et al., 2020)

| Type of reasoning | Percentage (%) |
|---|---|
| **Table → Passage** | 23.4 |
| **Passage → Table** | 20.3 |
| **Passage → Table →Passage** | 35.1 |
| **Passage and Table** | 17.3 |
| **Two parallel reasoning chain** | 3.1 |
| **multiple reasoning chains** | 0.8 |

**QASC** (Khot et al., 2020) is a multi-hop dataset that contains 9,980 8-way multiple-choice questions from elementary and middle school level science, and ach question is produced by composing two facts from an existing text corpus. To generate this dataset, crowd workers have first received only the first fact fS. Then, they have freely chosen the second fact fL from other available facts. Finally, the questions have been created by compromising these two facts. Thus, the models require commonsense reasoning for the composition of these facts to find the answer. You can see the number of questions, and unique fS and fL in Table 14. In contrast to OpenbookQA that which the number of facts to answer a question is unclear, in QASC, it is clear that two facts are sufficient to answer a question. The number of questions, fS and fL has been shown in Table 14. Also a comparison information among this dataset and previous multi-hop dataset has been shown in Table 15. As you can see QASC in addition to preparing supporting facts, annotated them too, also Decomposition is not evident.

Table 14: The number of the questions, unique fS and fL in QASC (Khot et al., 2020)

|  | Train | Dev | Test |
|---|---|---|---|
| **Number of questions** | 8,134 | 926 | 920 |
| **Number of unique fS** | 722 | 103 | 103 |
| **Number of unique fL** | 6,157 | 753 | 762 |

Table 15: Compression information (Khot et al., 2020)

| Property | CompWebQ | HotpotQA | MultiRC | OpenBookQA | WikiHop | QASC |
|---|---|---|---|---|---|---|
| **Supporting facts are available** | N | Y | Y | N | N | Y |
| **Supporting facts are annotated** | N | Y | Y | N | N | Y |
| **Decomposition is not evident** | N | N | Y | Y | Y | Y |

**MuSeRC** (Fenogenova et al., 2020) is the first multi-hop MRC Russian dataset. The questions in Russian Multi-Sentence Reading Comprehension (MuSeRC) dataset rely on multiple sentences and commonsense knowledge. It contains more than 900 paragraphs and 12,805 sentences across 5 different domains, namely: (1) elementary school texts, (2) news, (3) fiction stories, (4) fairy tales, and (5) summaries of TV series and books (Figure 8). In this dataset, the answers are not necessarily a span of the passage, and there may also be more than one answer to a question. Thus, the models should choose the best possible answer. Questions and answers have been created by crowd workers and, questions that can be answered with only one sentence have been removed to ensure that all questions are multi-hop. Although the number of the question in MuSeRC are less than MultiRC (Khashabi et al., 2018) but all of its questions are multi-hop (less than 60% question of MultiRC is multi-hop question), you can see some comparison information between MuSeRC and MultiRC in Table 16.

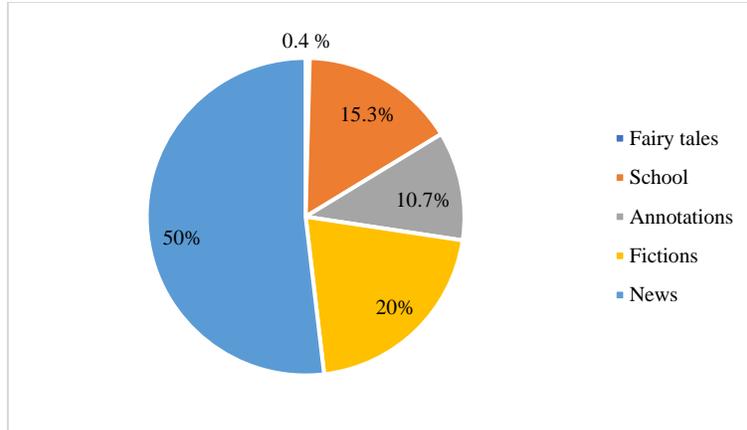

*Figure 8: The frequency of domains (Fenogenova et al., 2020)*

*Table 16: MuSeRC and MultiRC comparison (Fenogenova et al., 2020)*

| Property | MultiRC | MuSeRC |
|---|---|---|
| number of multi-hop questions | 5,825 | 5,228 |
| candidates / question | 5.28 | 4.16 |
| answers / question | 2.31 | 1.86 |
| sentence / passage | 14.3 | 13.875 |
| tokens / passage | 258.9 | 203.9 |
| tokens / question | 10.9 | 7.61 |
| tokens / answer | 4.7 | 5.3 |
| yes/no/true/false questions (%) | 27.57% | 1% |

**ComQA** (Wang et al., 2021) is a large-scale Chinese dataset that contains more than 120k human-labeled questions. They focus on compositional QA which means the answers have been obtained from multiple but discontinuous segments in the documents and the answers are not limited to one span but can be a combination of several separate sentences. Besides, as in this dataset HTML pages are used as the context, even a table or an image can be selected as a part of the answer. To generate the ComQA dataset, a question-document pair is obtained first from a Chinese search engine called Sogou Search[2], the content of the retrieved page is then converted into a list of sentences, and the crowd workers find the answer to the question through the context. In Figure 9, you can see the frequency and type of questions in this dataset.

There are different kinds of questions in this dataset: 1) Simple: questions about a single property of an entity. 2)Compositional: questions that require answering more primitive questions and combining them. 2)Temporal: questions that require temporal reasoning for deriving the answer. 3)Comparison: questions that need some kind of comparison. 4)Telegraphic: short questions in an informal manner similar to keyword queries. 4) Answer tuple: Where an answer is a tuple of connected entities as opposed to a single entity.  Table 17 shows Results of the manual analysis of 300 questions.

---

[2] https://www.sogou.com/

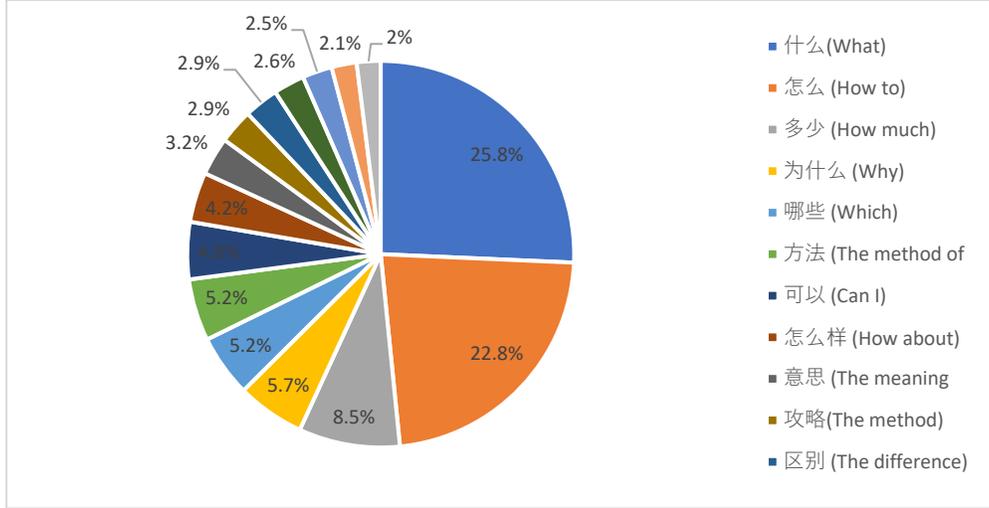

*Figure 9: The frequency and type of questions in ComQA (Wang et al., 2021)*

*Table 17: Types of questions in ComQA (Wang et al., 2021)*

| Property | Percentage% |
|---|---|
| Compositional questions | 32 |
| Temporal questions | 23.67 |
| Comparison questions | 29.33 |
| Telegraphic question | 8 |
| Answer tuple | 2 |

**MuSiQue** (Trivedi et al. 2022) is a multi-hop dataset via single-hop Question Composition. They first introduced a dataset construction approach that systematically selects composable pairs of single-hop questions that are connected and compose them to create multi-hop questions and then use this approach to build a multi-hop MRC dataset named MuSiQue. They ensure that no sub-question can be answered without finding the answer to the previous sub-questions it is connected to. This process of composing multi-hop questions chain can be seen in Table18. This dataset consists of 25K 2-4 hop questions with six different composition structure Table 19.

*Table 18: The six reasoning graph shapes (2-hop to 4-hop) present in MuSiQue(Trivedi et al. 2022)*

| Graph | Question | Decomposition |
|---|---|---|
| | Who succeeded the first President of Namibia? Hifikepunye Pohamba | 1. Who was the first President of Namibia? Sam Nujoma<br>2. Who succeeded Sam Nujoma? Hifikepunye Pohamba |
| | What currency is used where Billy Giles died? pound sterling | 1. At what location did Billy Giles die? Belfast<br>2. What part of the UK is Belfast located in? Northern Ireland<br>3. What is the unit of currency in Northern Ireland? pound sterling |
| | When was the first establishment that Mc-Donaldization is named after, open in the country Horndean is located? 1974 | 1. What is McDonaldization named after? McDonald's<br>2. Which state is Horndean located in? England<br>3. When did the first McDonald's open in England? 1974 |

| | When did Napoleon occupy the city where the mother of the woman who brought Louis XVI style to the court died? 1805 | 1. Who brought Louis XVI style to the court? Marie Antoinette
2. Who's mother of Marie Antoinette? Maria Theresa
3. In what city did Maria Theresa die? Vienna
4. When did Napoleon occupy Vienna? 1805 |
|---|---|---|
| | How many Germans live in the colonial holding in Aruba's continent that was governed by Prazeres's country? 5 million | 1. What continent is Aruba in? South America
2. What country is Prazeres? Portugal
3. Colonial holding in South America governed by Portugal? Brazil
4. How many Germans live in Brazil? 5 million |
| | When did the people who captured Malakoff come to the region where Philipsburg is located? 1625 | 1. What is Philipsburg capital of? Saint Martin
2. Saint Martin is located on what terrain feature? Caribbean
3. Who captured Malakoff? French
4. When did the French come to the Caribbean? 1625 |

*Table 19: Dataset statistics of MuSiQue(Trivedi et al. 2022)*

| | 2-hop | 3-hop | 4-hop | Total (24,814) |
|---|---|---|---|---|
| **Train** | 14376 | 4387 | 1175 | 19938 |
| **Dev** | 1252 | 760 | 405 | 19938 |
| **Test** | 1271 | 763 | 425 | 2459 |

For the last part of this section, Figure 10 has been prepared to summarize of the reviewed datasets. NewsQA, Race, and TriviaQA has known as multi-passages datasets but because of the multi-sentences question, they were reviewed here. The rest of the datasets has known as multi-hop datasets based on the problem definition in section 2.

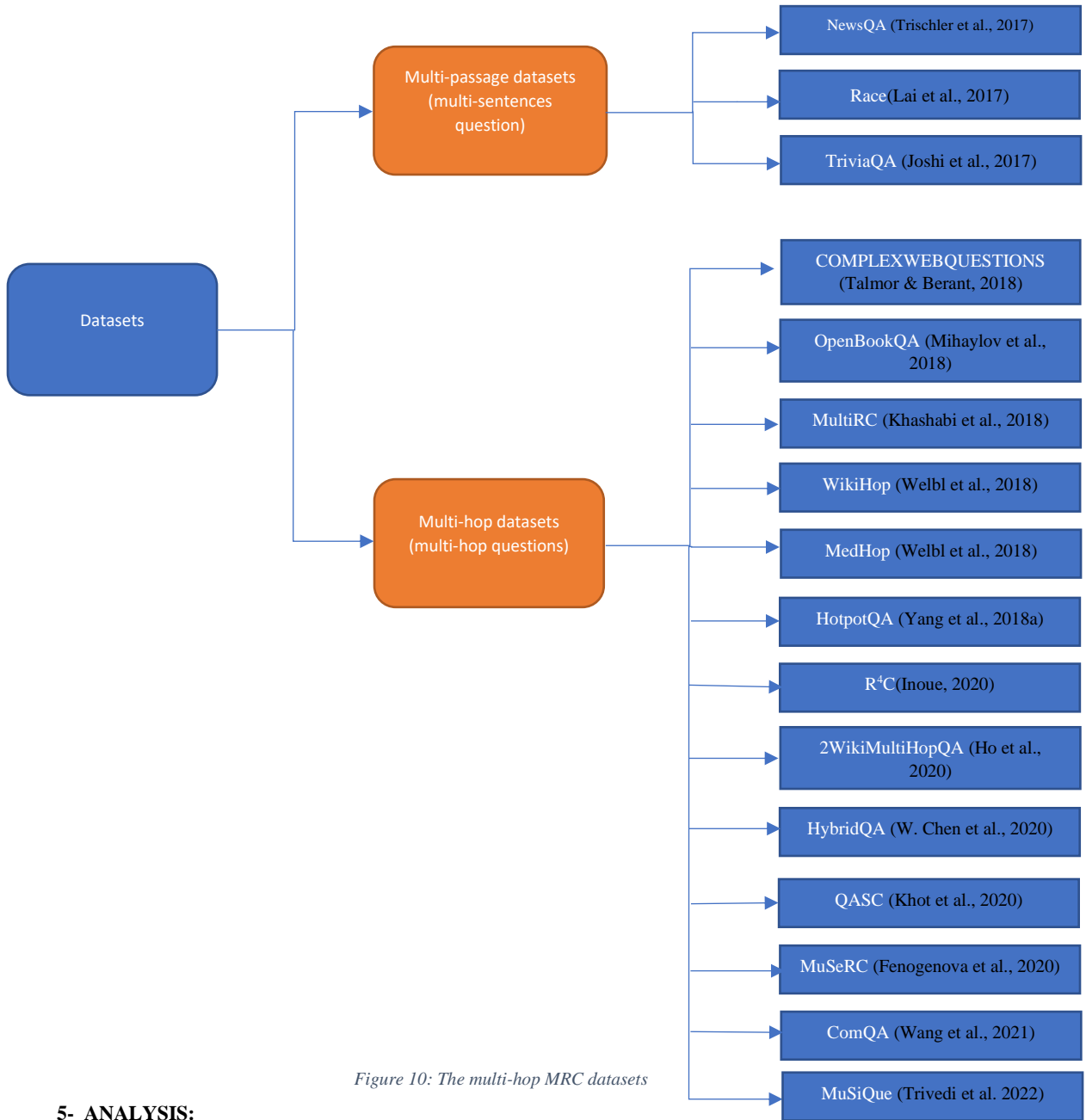

*Figure 10: The multi-hop MRC datasets*

## 5- ANALYSIS:

After reviewing each dataset in detail, some fine-grained comparisons will be presented in this section by preparing some figures and tables.

### 5-1 Frequency

First of all, the frequency usage of each dataset in recent multi-MRC models has been shown in Figure 11 (we used the same studies as Figure 2 which presented as Appendix1). As you can see, HotpotQA(Yang et al., 2018a) and WikiHop (Welbl et al., 2018) have received the most attention among the multi-hop datasets. There are several reasons for this:

1. They are considered pioneers in multi-hop MRC because They were the first successful and serious attempt to cover multi-hop challenges and they attract a great attention to the multi-hop MRC task.

2. Although some datasets were introduced recently, since these two datasets have been used by many multi-hop models, they provide the good situation for comparing the performance of the models. Therefore, models would like to use these two datasets to compare their performance with other models, and this leads to more use of these two datasets.

We also prepare Figure 10 to investigate the growth trends of HotpotQA and WikiHop from 2018 to 2021 (note that both datasets have been introduced in 2018). As you can see, HotpotQA is more popular than WikiHop, there are some reasons for it:

1. Questions in the WikiHop dataset are in triple form, while questions in the HotpotQA dataset are in the plain text form, which makes it more similar to real-world questions.
2. Answers in HotpotQA are in the form of Span-extraction while answers in WikiHop are multiple-choice. As mentioned in Section 2, the Span-extraction task is more popular among models so the frequency usage of HotpotQA will be more.
3. The HotpotQA dataset presents a new task called the Supporting fact task while WikiHop only focuses on the Answer task. So HotpotQA prepares a good situation to evaluate the ability of reasoning of the models.

Besides, as Figure 12 the number of usages of both datasets is maximum in 2019, which is due to the fact that the introduction of these two datasets in 2018 caused a lot of attention to the multi-hop MRC task, then in 2019, many multi-hop MRC models were proposed based on these two datasets. (It has been shown in Figure 2 too). In the following years, due to the introduction new multi-hop datasets, the usage of these two datasets got decreased but, both datasets are still the most popular datasets for multi-hop MRC models.

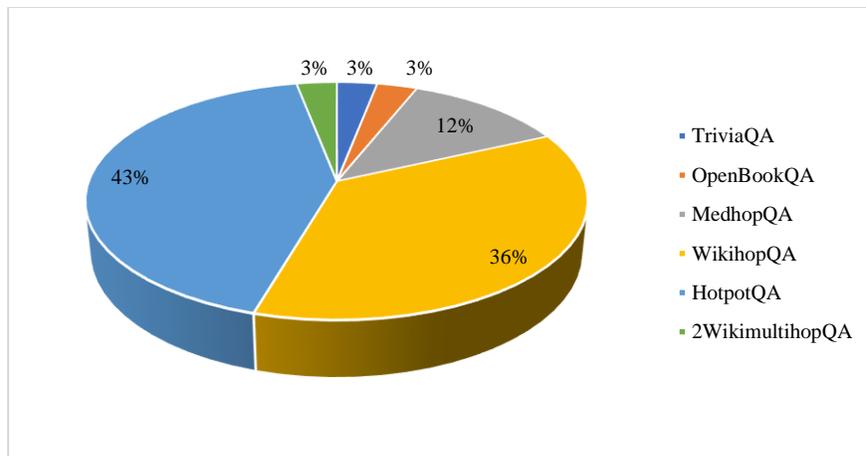

Figure 11: The frequency of each dataset

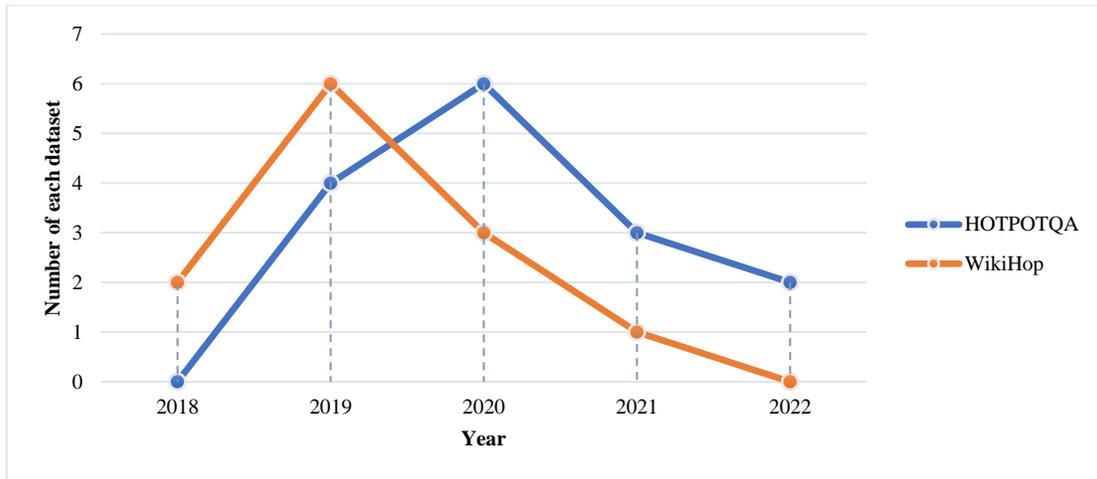

Figure 12: The HotpotQA and WikiHop growth trend

**5-2 Features**

Table 20 contains some composition information of features of the reviewed multi-hop MRC datasets. In this table, 15 datasets are investigated, including general features like Domain, Language, Task, Number of questions, Release date, Average length of passages, and the number of passages per question. This information helps readers to compare existing datasets.

Also, there is a feature about how many portions of each dataset are multi-hop. As it has been said before, we begin with some simple datasets that the questions can't be answered with a single sentence, although the sentences are within one passage and only a portion of these datasets include multi-sentence questions. These datasets are considered the first attempts to move from the single-hop datasets. (Note that some datasets are still used by recent multi-hop models, for example, DRNQA (X. Li et al., 2020) which has been published in 2020, has used TriviaQA (Joshi et al., 2017))

With emerging datasets that especially have focused on the multi-hop challenge (2018), in addition to preparing more complex questions, the portions of the multi-hop questions have been increased in datasets (in most cases, all the questions are datasets).

*Table 20: Statistical information of datasets*

| | Dataset | Domain | Language | Task | Number of questions | Release date | Average length of each passage | Number of passages per question | Portion of multi-hop questions % |
|---|---|---|---|---|---|---|---|---|---|
| 1) | NewsQA (Trischler et al., 2017) | Open | English | Span-extraction | 100K | 2017 | 30.7 sentences | 1 document | 20.7 |
| 2) | Race(Lai et al., 2017) | Open | English | Multiple-choice | 100K | 2017 | 321.9 tokens | 1 passage | 25.8 |
| 3) | TriviaQA (Joshi et al., 2017) | Open | English | Span-extraction | 95K | 2017 | 2,895 words | 6 documents | 40 |
| 4) | COMPLEXWEB QUESTIONS (Talmor & Berant, 2018) | Open | English | Multiple-choice | 43K | 2018 | 13.18 tokens | --- | 100 |
| 4) | OpenBookQA (Mihaylov et al., 2018) | Science facts | English | Multiple-choice | 6k | 2018 | 9.38 tokens | 1.16 facts | 100 |
| 5) | MultiRC (Khashabi et al., 2018) | Open | English | Multiple-choice | 9k | 2018 | 263.1 tokens | 1 paragraph | 60 |
| 6) | WikiHop (Welbl et al., 2018) | Open | English | Multiple-choice | 51k | 2018 | 100.4 tokens | 13.7 documents | 91 |
| 7) | MedHop (Welbl et al., 2018) | Molecular Biology | English | Multiple-choice | 2k | 2018 | 253.9 tokens | 36.4 documents | 100 |
| 8) | HotpotQA (Yang et al., 2018a) | Open | English | Span-extraction | 113K | 2018 | --- | 10 paragraphs | 100 |
| 9) | R$^4$C(Inoue, 2020) | Open | English | Multiple-choice | 4,588 | 2020 | --- | 10 paragraphs | 100 |
| 10) | 2WikiMultiHopQA (Ho et al., 2020) | Open | English | Span-extraction | 192K | 2020 | --- | 10 paragraphs | 100 |
| 11) | HybridQA (W. Chen et al., 2020) | Open | English | Span-extraction | 70K | 2020 | 15.7 rows, 4.4 columns | 1 table | 100 |
| 12) | QASC (Khot et al., 2020) | Science | English | Multiple-choice | 9K | 2020 | --- | 2 facts | 100 |
| 13) | MuSeRC (Fenogenova et al., 2020) | Open | Russian | Multiple-choice | 5K | 2020 | 1.5k characters | 1 passage | 100 |
| 14) | ComQA (Wang et al., 2021) | Open | Chinese | Span-extraction | 120K | 2021 | 581 words | 1 document | 100 |
| 15) | MuSiQue (Trivedi et al. 2022) | Open | English | Span-extraction | 24K | 2022 | 20-300 words | 10 paragraphs | 100 |

**5-3 Result**

As the last review of multi-hop datasets in this section, the stat-of-the-art result of existing multi-hop models on the reviewed multi-hop dataset will be shown. we considered the multi-hop MRC models from 2020 to 2021. As shown in Section X, the two MRC tasks, Span-extraction and Multiple-choice, have received the most attention among datasets, so we will review them separately.It should be noted that due to the different models, the results are not expressed to compare the performance of datasets, but only to show the state-of-the-result on each database.

*5-3-1 Span-extraction datasets:*

Since the answers are in the span-extraction form the EM and F1 are used as the evaluation metrics (Answer, Support, and Joint task). Table 21 shows the stat-of-the-result of the span-extraction dataset alongside the models. HotPotQA (Yang et al., 2018a) and 2WikiMultiHopQA (Ho et al., 2020) prepared all three task evaluation results while TriviaQA (Joshi et al., 2017), MuScRC (Fenogenova et al., 2020), and Hybrid (W. Chen et al., 2020) only focus on the Answer task.

*Table 21: The stat-of-the-result on span-extraction multi-hop datasets*

| Dataset | Model | Answer | | Support | | Joint | |
|---|---|---|---|---|---|---|---|
| | | EM | F1 score | EM | F1 score | EM | F1 score |
| HotpotQA-distractor (Yang et al., 2018a) | AMGN(R. Li et al., 2021) | 83.37 | 83.46 | 88.83 | 89.13 | 75.24 | 75.48 |
| HotPotQA-fullwiki (Yang et al., 2018a) | HGN (Fang et al., 2020) | 57.85 | 69.93 | 51.01 | 76.82 | 37.17 | 60.74 |
| 2WikiMultiHopQA (Ho et al., 2020) | RERC (Fu et al., 2021) | 71.56 | 74.51 | 86.00 | 92.75 | 50.59 | 60.21 |
| TriviaQa (Joshi et al., 2017) | DRNQA (X. Li et al., 2020) | 59.73 | 62.21 | | | | |
| MuSeRC (Fenogenova et al., 2020) | MuScRC (Fenogenova et al., 2020) | 25.6 | 65.6 | | | | |
| HybridQA (W. Chen et al., 2020) | HybridQA (W. Chen et al., 2020) | 43.8 | 50.6 | | | | |

*5-3-2 Multiple-choice datasets:*

Table 22 contains the stat-of-the-result of multi-hop models on multiple-choice datasets. Since the answer type of this dataset is multiple-choice then *Accuracy* is the evaluation metric on the test and development set.

*Table 22: The stat-of-the-result on multiple-choice multi-hop datasets*

| Dataset | Model | Accuracy (%) | |
|---|---|---|---|
| | | Test set | Dev set |
| WikiHop (Welbl et al., 2018) | ChainEX (J. Chen et al., 2020) | 76.5 | 72.2 |
| MedHop (Welbl et al., 2018) | EPaR(Jiang et al., 2019) | 60.3 | --- |
| OpenBookQA (Mihaylov et al., 2018) | PathNet(Kundu et al., 2018) | 53.4 | 55.00 |
| QASC (Khot et al., 2020) | MSSM(Q. Chen, 2022) | | 81.52 |

## 6- OPEN ISSUES

Datasets with the new challenges are a good motivation to improve current models. So, to reduce the gap between existing multi-hop MRC models and real-world cases, creating the datasets with more challenges is vital. Even the most popular multi-hop datasets still have shortcomings, then if models only focus on the existing dataset, the gap between them and real-world cannot be properly reduced, so it is important to focus on building datasets.

The lack of free and close-style datasets has made it impossible to present models for these two tasks, while both tasks have a good potential to use in form of multi-hop MRC.

Among all the types of MRC tasks described in Section 2, the free-form task is the most similar task to the real-world case but due to its complexity, there is no proper dataset for it. Since this type of answer is more similar to real-world scenarios, focusing on presenting free-form datasets to encourage models can improve the application of MRC systems.

## 7- CONCLUSION

In this study, a comprehensive survey on the multi-hop MRC evaluation metrics and dataset as two vital and important aspects of multi-hop MRC has been presented with a focus on the existing evaluation metrics and datasets, their growth trend and the important challenges of this area. In the beginning, the definition of the multi-hop MRC problem has been presented then the evaluation metrics have been investigated in detail. In following, 15 multi-hop datasets from 2017 to 2022 had been reviewed in detail then and some comprehensive analyses were presented for a fine-grain comparison of the different features of datasets. Finally, open issues in this field had been discussed.